\documentclass[runningheads]{llncs}

\usepackage{eccv}

\usepackage{eccvabbrv}

\usepackage{graphicx}
\usepackage{booktabs}

\usepackage[accsupp]{axessibility}  %

\usepackage{multirow}
\usepackage{pifont}
\usepackage{tabularray}
\usepackage{booktabs}
\usepackage[table]{xcolor} %
\usepackage{makecell}
\usepackage[breaklinks,colorlinks]{hyperref}
\usepackage{wrapfig}

\definecolor{highlightgray}{gray}{0.95} %
\definecolor{lightgold}{rgb}{0.99, 0.95, 0.85}

\usepackage{listings}
\usepackage{xcolor}

\definecolor{codegreen}{rgb}{0,0.6,0}
\definecolor{codegray}{rgb}{0.5,0.5,0.5}
\definecolor{codepurple}{rgb}{0.58,0,0.82}
\definecolor{backcolour}{rgb}{0.95,0.95,0.92}

\lstdefinestyle{mystyle}{
    backgroundcolor=\color{backcolour},   
    commentstyle=\color{codegreen},
    keywordstyle=\color{magenta},
    numberstyle=\tiny\color{codegray},
    stringstyle=\color{codepurple},
    basicstyle=\ttfamily\tiny,
    breakatwhitespace=false,         
    breaklines=true,                 
    captionpos=b,                    
    keepspaces=true,                 
    numbers=left,                    
    numbersep=5pt,                  
    showspaces=false,                
    showstringspaces=false,
    showtabs=false,                  
    tabsize=2
}

\usepackage{orcidlink}
\newcommand\blfootnote[1]{%
  \begingroup
  \renewcommand\thefootnote{}\footnote{#1}%
  \addtocounter{footnote}{-1}%
  \endgroup
}

\begin{document}

\title{GoldiCLIP: The Goldilocks Approach for Balancing Explicit Supervision for Language-Image Pretraining}

\titlerunning{GoldiCLIP}

\author{
    Deen Dayal Mohan\textsuperscript{*}%
    \and 
    Hossein Souri\textsuperscript{*}\and 
    Vitali Petsiuk\textsuperscript{*}\and
    \\
    Juhong Min\and
    Gopal Sharma\and
    Luowei Zhou\and
    Suren Kumar\textsuperscript{\dag}
}

\authorrunning{D.~Mohan et al.}

\institute{
    AI Center -- Mountain View, Samsung Electronics \\
    \email{\{d.mohan1, h.souri, v.petsiuk, j.min2, gopal.sharma, luowei.zhuo\}@samsung.com}
}

\maketitle
\blfootnote{\textsuperscript{*} Equal contribution, order chosen randomly.}
\blfootnote{\textsuperscript{\dag} Work done while at Samsung Electronics.}
\begin{abstract}
Until recently, the success of large-scale vision-language models (VLMs) has primarily relied on billion-sample datasets, posing a significant barrier to progress.
Latest works have begun to close this gap by improving supervision quality, but each addresses only a subset of the weaknesses in contrastive pretraining. We present GoldiCLIP, a framework built on a Goldilocks principle of finding the right balance of supervision signals. Our multifaceted training framework synergistically combines three key innovations: (1) a text-conditioned self-distillation method to align both text-agnostic and text-conditioned features; (2) an encoder integrated decoder with Visual Question Answering (VQA) objective that enables the encoder to generalize beyond the caption-like queries; and (3) an uncertainty-based weighting mechanism that automatically balances all heterogeneous losses.
Trained on just 30 million images, 300$\times$ less data than leading methods, GoldiCLIP achieves state-of-the-art among data-efficient approaches, improving over the best comparable baseline by 2.2 points on MSCOCO retrieval, 2.0 on fine-grained retrieval, and 5.9 on question-based retrieval, while remaining competitive with billion-scale models. Project page: \url{https://petsi.uk/goldiclip}.
\end{abstract}

\section{Introduction}
\label{sec:intro}

Contrastively trained vision-language encoders like CLIP~\cite{radford2021learning} have become foundational components for numerous vision tasks and serve as the backbone for integrating visual understanding into large language models~\cite{liu2023visual}. Following the scaling paradigm established in NLP, the dominant approach to improving these models has been to increase dataset size. CLIP pioneered this direction with 400 million image-text pairs, while recent state-of-the-art models such as SigLIP~2~\cite{tschannen2025siglip2} and Perception Encoder~\cite{bolya2025perception} leverage proprietary datasets of 5--10 billion images. Such data requirements pose a significant barrier, restricting progress to well-resourced industrial laboratories.

Recent works have begun to challenge this paradigm by demonstrating that higher-quality data\cite{zheng2024dreamlip,bulat2024fff} and supervision can partially compensate for limited data. FLAIR~\cite{xiao2025flair} introduced the text-conditioned visual representations via cross-attention pooling, improving fine-grained retrieval. COSMOS~\cite{kim2025cosmos} incorporated cross-modal self-distillation to enhance spatial understanding. These methods have shown that models trained on 30 million images can approach the retrieval performance of billion-scale counterparts. 

Designed individually, these supervision objectives are studied in isolation or integrated via rigid training pipelines (\eg, staged training~\cite{tschannen2025siglip2}). Effectively combining these disparate signals remains a non-trivial challenge: different losses often operate at different scales, target conflicting feature granularities, and require extensive hyperparameter tuning to prevent one signal from dominating or degrading the others. Consequently, the potential synergy between these complementary forms of supervision remains largely untapped.

In this work, we present GoldiCLIP, a unified training framework
for data-efficient vision-language pretraining. The design principle behind GoldiCLIP is that supervision signals should not be treated as independent additions to a contrastive baseline. Instead, each component is carefully adapted to the overall architecture: we extend self-distillation to operate on text-conditioned features rather than only text-agnostic ones, route decoder tasks through the text encoder to improve its generalization beyond caption-like queries, and employ learned task balancing to ensure stable optimization across all objectives jointly. Our experiments demonstrate that these modifications have a positive contribution to the overall performance.
Our key contributions can be summarized as follows:
\begin{itemize}
\item \textbf{Text-Conditioned Local-Global Distillation:} We implement a novel self-distillation mechanism that establishes correspondences between local views and global representations while accounting for the textual context. This ensures that the learned visual features are not only spatially consistent but also semantically relevant to the associated text.
\item \textbf{Encoder-integrated decoder with VQA supervision:} We first extend \cite{wan2024locca} decoder by incorporating VQA objective. Further we route all decoder objectives through the text encoder, encouraging the encoder to generalize beyond the caption-like queries, a common failure mode in models trained using re-captioned dataset.

\item \textbf{Automated Supervision Balancing:} To avoid the complexities of manual weight tuning or multi-stage training schedules, we frame the integration as a multi-task learning problem. We employ uncertainty-based weighting \cite{kendall2018multi} to automatically balance the contribution of each loss in real-time, ensuring stable optimization and seamless signal fusion.
\end{itemize}

We evaluate GoldiCLIP extensively across six task categories: standard image-text retrieval, fine-grained-retrieval, long-text retrieval, question-based retrieval, zero-shot semantic segmentation and zero-shot classification. We further provide detailed ablations showing the contribution of the each components. Our framework is trained entirely on open-source data, providing a reproducible recipe for the research community. Our results demonstrate that we achieve state-of-art results across models trained on similar scale of data.
On standard text-to-image retrieval benchmark, MSCOCO \cite{lin2014microsoft}, GoldiCLIP trained on just 30 million unique images, outperforms FLAIR by 2.2 and COSMOS by 3 percentage points (pp) on text-to-image retrieval. The gains are even more pronounced on fine-grained retrieval tasks. On the IIW-FG\cite{garg2024imageinwords} benchmark, we improve over FLAIR and COSMOS by 2.9pp and 3.9pp respectively. Even when trained on only 30 Million images, our method improves over prior state-of-the-art models such as SigLIP 2, Perception Encoder etc., which are trained on large scale data (5-10 billion image) on many of these benchmarks. 

\section{Related Work}
\label{sec:related_work}

Akin to LLM research, the field of large-scale vision-language pre-training has been dominated by the paradigm of scaling datasets and model sizes to improve performance. CLIP \cite{radford2021learning} pioneered this approach with a dual-encoder architecture trained contrastively on 400 million images paired with text, demonstrating remarkable zero-shot transfer cap
abilities. ALIGN \cite{jia2021scaling} scaled this to over 1 billion images, while more recent models such as SigLIP 2 \cite{tschannen2025siglip2} and Perception Encoder \cite{bolya2025perception} and MetaCLIP \cite{xu2023demystifying} leverage 2-10 billion images with paired text to achieve state-of-the-art performance. Eva-CLIP \cite{sun2024eva} explored the second dimension of scaling, reaching 18 billion model parameters. However, both scaling approaches, dataset size and model parameters, remain inaccessible to most researchers due to computational and storage requirements.

\textbf{Auxiliary Supervision Objectives.}
Recognizing the limitations of single global vector-based contrastive learning, researchers have proposed various auxiliary objectives to enhance vision-language models. Generative decoders represent one major direction. Inspired by image captioning methods like GiT \cite{wang2022git} and LEMON \cite{hu2022scaling}, CoCa \cite{yu2022coca} unified text generation with contrastive learning. LocCa \cite{wan2024locca} introduced multi-task decoder objectives including captioning, referring expressions, and grounded captioning to provide more granular supervision. SigLIP 2 \cite{tschannen2025siglip2} further extended this approach, combining multi-task decoders with self-supervised losses including self-distillation and masked prediction. However, these methods have been primarily evaluated in data-rich settings, leaving their effectiveness in data-limited regimes unexplored.

\textbf{Self-distillation techniques} represent another important category of auxiliary supervision. Drawing inspiration from self-supervised learning methods like DINO \cite{caron2021emerging}, SLIP \cite{mu2022slip} integrated self-supervised learning (SSL) approaches into CLIP to improve transferability. SILC \cite{naeem2024silc} applied self-distillation to establish better local-global correspondences within image representations, using an exponential moving average (EMA) teacher to distill knowledge into a student encoder processing different views of the same image. TIPS \cite{maninis2024tips} combined self-distillation with masked image modeling to encourage spatial coherence. Cross-attention pooling between image and text features to improve fine-grained alignment was introduced by \cite{xiao2025flair}. The pooling creates a text-conditioned visual representations that adapt to specific queries. A concurrent work \cite{kim2025cosmos} incorporated similar pooling mechanism into the self-supervision pipeline. However, a critical limitation of existing self-distillation methods is that they do not explicitly supervise text-conditioned representations that have proven effective for fine-grained retrieval.

\textbf{Loss Balancing and Multi-Task Learning.}
As models incorporate multiple auxiliary objectives, balancing the contribution of different loss terms becomes a significant challenge. SigLIP 2 \cite{tschannen2025siglip2} manually chose both loss weights and training regimes (e.g., adding SSL losses at 80\% of training completion), while CoCa \cite{yu2022coca} relied on grid search of hyperparameters. 
In the multi-task learning literature, Uncertainty-based weighting \cite{kendall2018multi} proposes learning task-specific uncertainty parameters that automatically balance losses during training, avoiding expensive hyperparameter search. 
To the best of our knowledge, we are the first work to integrate this approach in application to vision-language pretraining, which helps mitigate the difference in scales (token-level generation vs. global contrastive matching).

\textbf{Synthetic Caption Generation.}
An alternative approach to reducing data requirements focuses on augmenting existing datasets with caption augmentation leveraging LLMs and VLMs. \cite{fan2023improving, bulat2024fff, zheng2024dreamlip, zhang2024long, wu2024lotlip, patel2024tripletclip, xie2025fg}. Generation of extra captions is considered text-based data augmentation, as set of images used for training remains constant.  While these approaches have proven effective for caption-like queries, performance degrades significantly when queries require deeper reasoning such as questions about specific objects, spatial relationships, or attributes \cite{zheng2024dreamlip}. This limitation suggests that diversifying the \textit{types} of supervision, not just augmenting captions, may be necessary to close the performance gap.

\section{Method}
\label{sec:method}

\begin{figure*}[t]
    \centering
    \includegraphics[width=\textwidth]{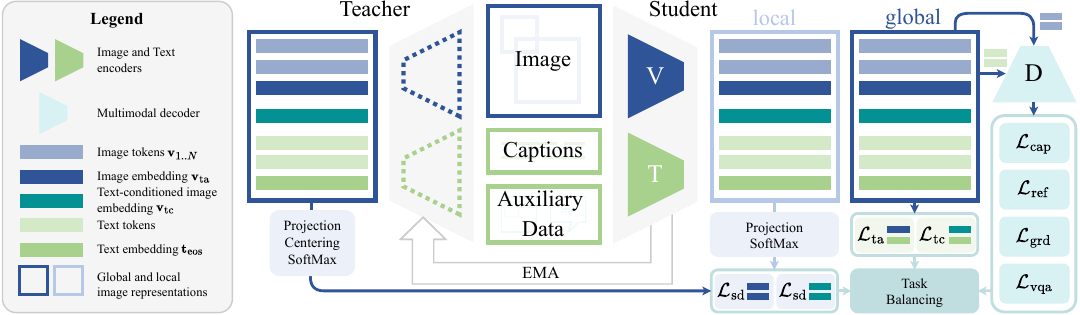}
    \caption{
        An overview of GoldiCLIP approach.
        Input images, captions, and auxiliary textual data (such as VQA) are processed by the student and teacher models, which consist of a text encoder and a vision encoder each. The teacher model is constructed as an EMA of a student, and is used to produce robust and stable global image representations which serve as a target for the local representations of the student model following the self-distillation approach (Sec \ref{subsec:selfdistillation}). 
        In our framework, the text embedding is contrastively aligned not only with the standard image embedding but also with the text-conditioned image embedding (Sec \ref{subsec:contrastive_objectives}). Multimodal decoder uses image patch tokens along with the textual tokens from the text encoder to perform generative tasks (such as VQA) (Sec~\ref{subsec:decoder_objectives}). Finally, all objectives are automatically weighted using a task-balancing approach (Sec \ref{subsec:task_balancing}).
    }
    \label{fig:architecture}
\end{figure*}

Motivated by the gaps identified in Section~\ref{sec:related_work}, we propose GoldiCLIP, a unified training framework that systematically integrates diverse supervision signals to dramatically improve data efficiency in vision-language pre-training. 
Figure~\ref{fig:architecture} provides a holistic overview of all components of our approach. 

We begin by describing our baseline architecture inspired by FLAIR~\cite{xiao2025flair} in Section~\ref{subsec:contrastive_objectives}. Section~\ref{subsec:selfdistillation} presents our \textbf{text-conditioned self-distillation method}, which differs from existing approaches~\cite{naeem2024silc} by operating on both text-agnostic and text-conditioned visual representations. 
Section \ref{subsec:decoder_objectives} introduces our enhanced multi-task decoder framework that extends prior work \cite{wan2024locca, yu2022coca, tschannen2025siglip2} with a \textbf{VQA objective}, which substantially improves performance on dense prediction tasks (\eg, semantic segmentation) and question-based retrieval.

Finally, Section~\ref{subsec:task_balancing} describes our approach to automatically balancing our diverse set of training losses using uncertainty-based multi-task learning, eliminating the need for expensive grid-search hyperparameter tuning.
\subsection{Model Design}
\label{subsec:contrastive_objectives}
Following the CLIP architecture~\cite{radford2021learning}, our framework consists of a ViT-based visual encoder $V_{\theta}$ that processes an image $I$ into a global embedding $\mathbf{v}_\textrm{cls}$ and a sequence of patch tokens, and a textual encoder $T_{\theta}$ that processes text $T$ into a sentence embedding token $\mathbf{t}_\textrm{eos}$:
\begin{align*}
    V_{\theta}(I) &= [\mathbf{v}_\textrm{cls}, \mathbf{v}_1, \ldots, \mathbf{v}_N] \in \mathbb{R}^{(N + 1) \times \textrm{dim}_v} \\
    T_{\theta}(T) &= \mathbf{t}_\textrm{eos} \in \mathbb{R}^{\textrm{dim}_t}
\end{align*}

The design of the final visual feature used for contrastive alignment with $\mathbf{t}_\textrm{eos}$ critically impacts the learned representations. Traditional methods use the global image representation $\mathbf{v}_\textrm{cls}$~\cite{radford2021learning, zhai2023sigmoid}, which produces a fixed embedding efficient for retrieval but requires massive datasets to capture information relevant to diverse queries.
FLAIR~\cite{xiao2025flair} proposed cross-attention pooling, where the visual representation is explicitly conditioned on text features $\mathbf{t}_\textrm{eos}$:
\begin{equation}
    \mathbf{v}_\textrm{tc} = \omega({\mathbf{v}_1, \ldots, \mathbf{v}_N}; \mathbf{t}_\textrm{eos}),
\end{equation}
where $\omega$ denotes a cross-attention operation with the text feature as query and projections of visual tokens as keys and values. 

We employ a sigmoid-based contrastive loss from FLAIR~\cite{xiao2025flair}, denoted as $\mathcal{L}_\textrm{ret}$, computed as a sum of two terms comparing text representations with both text-conditioned and text-agnostic visual representations ($\mathbf{v}_\textrm{ta}=\mathbf{v}_\textrm{cls}$).
Following the text augmentation strategy described in \cite{xiao2025flair}, we sample $K$ captions for each of the $B$ images in a batch, yielding $B \times (B \times K)$ image-caption pairs.
For each pair of the $i$-th image and $(j, k)$-th caption ($i,j\in\{1,\ldots,B\}, k\in\{1,\ldots,K\}$), we define $y_{i,j,k} = +1$ if the caption corresponds to the image ($j=i$) and $-1$ otherwise.
Denoting image representations as $\mathbf{v}_\textrm{ta}^{i}$ and $\mathbf{v}_\textrm{tc}^{i,j,k}$, and text embeddings as $\mathbf{t}^{j,k}$, the loss for this pair is:
\begin{align}
    \mathcal{L}_\textrm{ret}^{i,j,k} &= \mathcal{L}_\textrm{tc}^{i,j,k} + \mathcal{L}_\textrm{ta}^{i,j,k} \nonumber \\
    &= \frac{1}{1 + \exp(y_{i,j,k}(-t\langle\mathbf{v}_\textrm{tc}^{i,j,k}, \mathbf{t}^{j,k}\rangle + b))} \label{eq:ret_loss} \\
    &+ \frac{1}{1 + \exp(y_{i,j,k}(-t\langle\mathbf{v}_\textrm{ta}^{i}, \mathbf{t}^{j,k}\rangle + b))}, \nonumber
\end{align}
where learnable parameters $t$ and $b$ represent temperature and bias, and $\langle \cdot, \cdot\rangle$ denotes cosine similarity.

\subsection{Text-Conditioned Self-Distillation}
\label{subsec:selfdistillation}
Improving local-to-global consistency enhances vision encoder performance on downstream applications, especially dense prediction tasks~\cite{naeem2024silc}. Self-distillation techniques like SILC~\cite{naeem2024silc} and TIPS~\cite{maninis2024tips} establish this consistency by distilling knowledge from a teacher to a student model. However, these methods operate exclusively on text-agnostic visual features (typically the \texttt{[CLS]} token), ignoring text-conditioned representations that have proven effective for fine-grained retrieval~\cite{xiao2025flair}.\\
\indent\textbf{Our Approach.}
We propose a novel self-distillation method that extends to both text-conditioned and text-agnostic representations.
We maintain the teacher model's weights $\theta^T$ as an exponential moving average (EMA) of the student's weights $\theta^S$ with a momentum $m_\textrm{sd}$. \\
\indent To introduce local-to-global consistency, we generate multiple views of each image $I$: local views (smaller random crops) and global views (larger crops). The teacher processes only global views, providing stable global-level knowledge targets. The student processes both local and global views and learns to align its local representations with the teacher's global ones. Specifically for batch size $B$, with $K'$ as the total number of captions in the batch (positive and negative) used to condition visual representations, and $N_\textrm{glob}$ and $N_\textrm{loc}$ as the local and global views, the text-conditioned representation $\mathbf{v}_\textrm{tc}^T$ and text-agnostic representation $\mathbf{v}_\textrm{ta}^T$ from the teacher have shapes of $B \times K' \times N_\textrm{glob} \times \textrm{dim}_v$ and $B \times N_\textrm{glob} \times \textrm{dim}_v$, respectively. The representations $\mathbf{v}_\textrm{tc}^S$ and $\mathbf{v}_\textrm{ta}^S$ from the student have the same shapes except for the number of views $N_{loc}$ instead of $N_{glob}$.

We linearly transform student and teacher features into a shared space using projection matrices $\mathbf{W}^S$ and $\mathbf{W}^T$:
\begin{align}
    \mathbf{y}^{T}_{*} &= \mathbf{W}^T \mathbf{v}^{T}_{*}, \quad
    \mathbf{y}^{S}_{*} = \mathbf{W}^S \mathbf{v}^{S}_{*},
\end{align}
where $*$ denotes all applicable indices.
For the teacher, the projected features are 
centered and scaled by the learned temperature $\tau_T$, for the student they are scaled by temperature $\tau_S$:
\begin{align}
    \mathbf{z}^{T}_\textrm{ta} &= (\mathbf{y}^{T}_\textrm{ta} - \mathbf{c}_\textrm{ta}) / \tau_T, \\
    \mathbf{z}^{T}_\textrm{tc} &= \left(\frac{1}{K'} \sum_{k=1}^{K'}\left[\mathbf{y}^{T}_\textrm{tc}\right]_k - \mathbf{c}_\textrm{tc}\right) / \tau_T \\
    \mathbf{z}^{S}_\textrm{ta} &= \mathbf{y}^{S}_\textrm{ta} / \tau_S, \\
    \mathbf{z}^{S}_\textrm{tc} &= \left(\frac{1}{K'} \sum_{k=1}^{K'}\left[\mathbf{y}^{S}_\textrm{tc}\right]_k \right) / \tau_S
\end{align}
Note that for text-conditioned representations, we average over all $K'$ captions before applying centering and temperature scaling. This is an important design choice: a local view captures only a small crop of the image and may contain little information relevant to a given caption. Conditioning these local views on individual captions can therefore produce noisy representations. Directly aligning these representations to the teacher's global text-conditioned representations risks training instability or collapse. Averaging over captions smooths out this noise and ensures stable distillation dynamics.

The logits $\mathbf{z}$ are transformed into probability distributions $p$ using SoftMax. The student is trained by minimizing the distillation loss $\mathcal{L}_\textrm{sd}$, which is a sum of cross-entropy terms between teacher target distributions and student predicted distributions across different views:
\begin{align}
    \mathcal{L}_\textrm{sd} = \sum_{i=1}^{N_\textrm{glob}} \sum_{j=1}^{N_\textrm{loc}} H\left(p^{T,i}_\textrm{ta}, p^{S,j}_\textrm{ta}\right) + H\left(p^{T,i}_\textrm{tc}, p^{S,j}_\textrm{tc}\right),
\end{align}
where $H$ is a cross-entropy function.

The centers $\mathbf{c}_\textrm{ta}$ and $\mathbf{c}_\textrm{tc}$ are updated with EMA momentum $m_c$ using batch means of the projected teacher features. We provide detailed formulation in supplementary material.

\subsection{Decoder Objectives}
\label{subsec:decoder_objectives}
Prior work~\cite{tschannen2025siglip2, yu2022coca, wan2024locca} demonstrates that incorporating generative objectives alongside contrastive learning produces more robust visual representations. 
Following~\cite{yu2022coca,wan2024locca}, we introduce a transformer-based decoder $D_\theta$ that autoregressively generates text sequences conditioned on visual features from the image encoder.
The textual prompt dictates one of multiple generation tasks proposed in~\cite{wan2024locca}: image captioning, referring expression generation, or grounded captioning.
For instance, for grounded captioning, the decoder receives a prompt ``\texttt{OBGR [}$x_1, y_1, x_2, y_2$\texttt{], grounded caption is}'' --- a task identifier with bounding-box coordinates, and generates a caption describing that object region.

\textbf{VQA Objective.} 
To further enrich the learned visual representations, we propose incorporating a Visual Question Answering (VQA) task into the decoder training framework. While descriptive objectives like caption generation are effective, VQA provides a more rigorous training signal that directly enhances the underlying visual features. The compositional reasoning required to answer complex questions forces the model to learn a more robust feature space, where object identities, attributes, and spatial relationships are better encoded --- capabilities not strictly required for caption generation alone.
We generate auxiliary VQA data using open-world detector models~\cite{minderer2023scaling} and large-scale VLMs.

The loss for each of the four decoder tasks is a standard negative log-likelihood for next token prediction, where the model learns to predict the ground-truth text sequence $\mathbf{T}$ given the task prompt and conditioned on the visual token embeddings $\mathbf{V}=[\mathbf{v}_1, \ldots, \mathbf{v}_N]$:
\begin{align}
\mathcal{L}_\textrm{task}(\theta) &= - \sum \log P_\theta(\mathbf{T}^\textrm{task}_{j} \mid \mathbf{T}^\textrm{task}_{<j}, \mathbf{V}),\\
    \textrm{task} &\in \{\textrm{cap, ref, grd, vqa}\}. \nonumber  
\end{align}

\textbf{Architecture Choice.}
Unlike LocCa~\cite{wan2024locca} and SigLIP 2 \cite{tschannen2025siglip2}, which feed tokenized text directly to the decoder $D_\theta$, we adopt a CoCa-style~\cite{yu2022coca} architecture where text is tokenized and then embedded using the text encoder $T_\theta$.
This design optimizes the text encoder on both the contrastive loss 
and all decoder objectives
, encouraging it to generalize beyond caption-like queries and handle diverse textual structures.
This generalization is particularly valuable when the model is integrated into retrieval-augmented vision-language systems with varied query distributions.
By eliminating the additional tokenizer, this also reduces the number of model parameters by 10\%, enabling larger batch sizes in training.
Our text encoder employs causal self-attention to prevent the decoder from accessing future tokens, thereby avoiding information leakage.

\subsection{Task Balancing}
\label{subsec:task_balancing}
Our model is optimized using a diverse set of six supervisory signals: contrastive retrieval ($\mathcal{L}_\textrm{ret}$), four generative decoder tasks ($\mathcal{L}_\textrm{cap}$, $\mathcal{L}_\textrm{ref}$, $\mathcal{L}_\textrm{grd}$, $\mathcal{L}_\textrm{vqa}$), and self-distillation ($\mathcal{L}_\textrm{sd}$). Balancing these heterogeneous signals is crucial but manually tuning loss weights is computationally prohibitive.

We frame training as a multi-task learning problem and employ uncertainty-based weighting~\cite{kendall2018multi}. Each loss term $\mathcal{L}_\textrm{task}$ is associated with a learnable uncertainty parameter $\sigma_{task}^2$ that represents task-dependent uncertainty. This allows the model to automatically down-weight tasks with higher uncertainty, balancing their influence on gradient updates. The final training objective jointly minimizes model weights and uncertainty parameters:
\begin{align}
    \mathcal{L}_\textrm{total} &= \sum_{\textrm{task}} \frac{\mathcal{L}_\textrm{task}}{\sigma_{\textrm{task}}^2} + \sigma_{\textrm{task}}^2,\\
    \textrm{task} &\in \{\textrm{ret, cap, ref, grd, vqa, sd}\}. \nonumber    
\end{align}
The $\sigma_{\textrm{task}}^2$ term prevents the trivial solution of infinite uncertainty. This principled approach eliminates expensive hyperparameter search and enables stable, efficient optimization across diverse supervision signals operating at vastly different scales (e.g., token-level generation vs.\ global contrastive matching).

\section{Experiments}
\label{sec:expts}

\begin{table*}[!t]
    \hypertarget{tab:retrieval_full}{} 
    \caption{Zero-shot image-text retrieval on \textit{(a) standard benchmarks} (validation splits of MSCOCO~\cite{lin2014microsoft}, Flickr30k~\cite{plummer2015flickr30k}), \textit{(b) fine-grained} retrieval setting introduced by \cite{xiao2025flair} (sentence-level on DOCCI~\cite{onoe2024docci}, IIW~\cite{garg2024imageinwords}), and \textit{(c) long-text} retrieval setting (DCI~\cite{urbanek2024picture}, Urban 1k~\cite{zhang2024long}).}
    \resizebox{\textwidth}{!}{
        \begin{tabular}{lc|cccccccc|cccc|cccc}
            & & \multicolumn{8}{|c|}{(a) Standard Benchmarks} & \multicolumn{4}{|c|}{(b) Fine-Grained} & \multicolumn{4}{|c}{(c) Long Text} \\
            \toprule
            \multirow{3}{*}{Method} & \multirow{3}{*}{\makecell{Data\\size}} & \multicolumn{4}{c}{MSCOCO} & \multicolumn{4}{c|}{Flickr30k} & \multicolumn{2}{c}{DOCCI-FG} & \multicolumn{2}{c|}{IIW-FG} & \multicolumn{2}{c}{DCI} & \multicolumn{2}{c}{Urban 1k} \\
            \cmidrule(lr){3-6}
            \cmidrule(lr){7-10}
            \cmidrule(lr){11-12}
            \cmidrule(lr){13-14}
            \cmidrule(lr){15-16}
            \cmidrule(lr){17-18}
            & & \multicolumn{2}{c}{T2I} & \multicolumn{2}{c}{I2T} & \multicolumn{2}{c}{T2I} & \multicolumn{2}{c|}{I2T} & T2I & I2T & T2I & I2T & T2I & I2T & T2I & I2T \\
            & & R1 & R5 & R1 & R5 & R1 & R5 & R1 & R5 & R1 & R1 & R1 & R1 & R1 & R1 & R1 & R1 \\
            \toprule
            DreamLIP~\cite{zheng2024dreamlip} & 3M & 29.8 & 55.4 & 40.8 & 68.4 & 53.6 & 78.4 & 69.2 & 91.5 & 10.3 & 23.3 & 22.7 & 59.2 & - & - & - & - \\
            FLAIR~\cite{xiao2025flair} & 3M & 37.7 & 64.5 & 51.6 & 77.2 & 65.7 & 86.8 & 78.7 & 95.2 & 15.2 & 35.7 & 30.5 & 70.6 & 50.3 & 47.3 & 69.5 & 63.5\\ 
            COSMOS~\cite{kim2025cosmos} & 3M & 40.1 & 66.8 & 53.1 & 78.3 & 68.6 & 88.7 & 84.1 & 96.5 & 14.2 & 32.9 & 28.9 & 70.9 & 52.5 & 50.2 & 73.2 & 67.1 \\ 
            \rowcolor{lightgold}
            GoldiCLIP & 3M & {45.1} & {71.2} & {58.6} & {82.4} & {73.8} & {91.2} & {88.7} & {97.4} & {18.7} & {43.0} & {35.4} & {79.4} & 57.7 & 52.8 & 79.3 & 72.3 \\ 
            \midrule
            SigLIP~\cite{zhai2023sigmoid} & 30M & 49.0 & 74.5 & 64.1 & 86.2 & 76.7 & 93.2 & 90.5 & 98.3 & - & - & - & - & - & - & - & - \\
            DreamLIP ~\cite{zheng2024dreamlip} & 30M & 44.8 & 69.8 & 62.3 & 84.5 & 73.3 & 91.8 & 89.9 & 99.0 & 21.6 & 51.2 & 37.5 & 85.3 & - & - & - & - \\
            COSMOS ~\cite{kim2025cosmos} & 30M & 52.5 & 77.2 & 68.0 & 87.8 & 80.3 & 95.3 & 92.9 & \textbf{99.4} & 23.1 & 57.3 & 40.1 & 90.2 & 67.7 & 63.7 & 88.0 & 82.7 \\
            FLAIR ~\cite{xiao2025flair} & 30M & 53.3 & 77.5 & 68.0 & 87.8 & 81.1 & 94.9 & 94.7 & 99.3 & 25.0 & 59.0 & 41.7 & 91.5 & 66.3 & 61.3 & 87.7 & \textbf{83.6}\\
            \rowcolor{lightgold}
            GoldiCLIP & 30M & \textbf{55.5} & \textbf{78.9} & \textbf{70.3} & \textbf{89.3} & \textbf{83.0} &\textbf{95.9} & \textbf{94.8} & 99.1 & \textbf{27.0} & \textbf{60.2} & \textbf{44.0} & \textbf{90.9} & \textbf{71.9} & \textbf{67.5} & \textbf{88.6} & 83.2 \\
            \midrule
            LiT~\cite{zhai2022lit} & 100M &  32.7 & 57.8 & 51.9 & 75.7 & 60.9 & 84.6 & 80.0 & 95.9 & - & - & - & - & 40.9 & 41.7 & - & - \\
            LoTLIP~\cite{wu2024lotlip} & 100M & 38.1 & 63.8 & 59.7 & 81.5 &  65.2 & 88.0 & 86.9 & 97.8 & - & - & - & - & 61.0 & 62.1 & - & - \\
            Long-CLIP~\cite{zhang2024long} & 400M & 40.3 & 65.9 & 57.3 & 80.8 & 70.7 & 90.6 & 85.9 & 98.5 & - & - & - & - & 44.1 & 47.4 & 79.5 & 78.9 \\
            OpenCLIP~\cite{cherti2023reproducible} & 2B & 41.7 & 67.1 & 59.3 & 82.4 & 71.9 & 90.4 & 87.5 & 97.7 & 17.4 & 49.7 & 30.6 & 84.1 & 55.4 & 56.0 & 65.8 & 69.5 \\
            PE$_\textrm{core}$~\cite{bolya2025perception} & 5B & 49.9 & 74.8 & \textbf{70.3} & 88.9 & 81.3 & 95.4 & 93.5 & 99.0 & 22.4 & 61.8 & 36.8 & 90.7 & 60.0 & 60.2 & 56.7 & 60.3 \\
            SigLIP~\cite{zhai2023sigmoid} & 10B & 47.2 & 72.1 & 65.5 & 86.2 & 75.6 & 92.8 & 89.1 & 98.6 & 20.6 & 57.5 & 33.8 & 83.7 & 56.2 & 57.7 & 62.1 & 62.7 \\
            SigLIP 2~\cite{tschannen2025siglip2} & 10B & 52.5 & 76.6 & 70.0 & 88.3 & 80.0 & 94.5 & 91.8 & 98.8 & 23.2 & 61.2 & 37.0 & 91.3 & 55.7 & 55.7 & 57.8 & 60.3 \\
            \bottomrule
        \end{tabular}
    }
    \label{tab:retrieval_full}
\end{table*}

\begin{figure}[!t]
    \centering
    \includegraphics[width=0.97\linewidth]{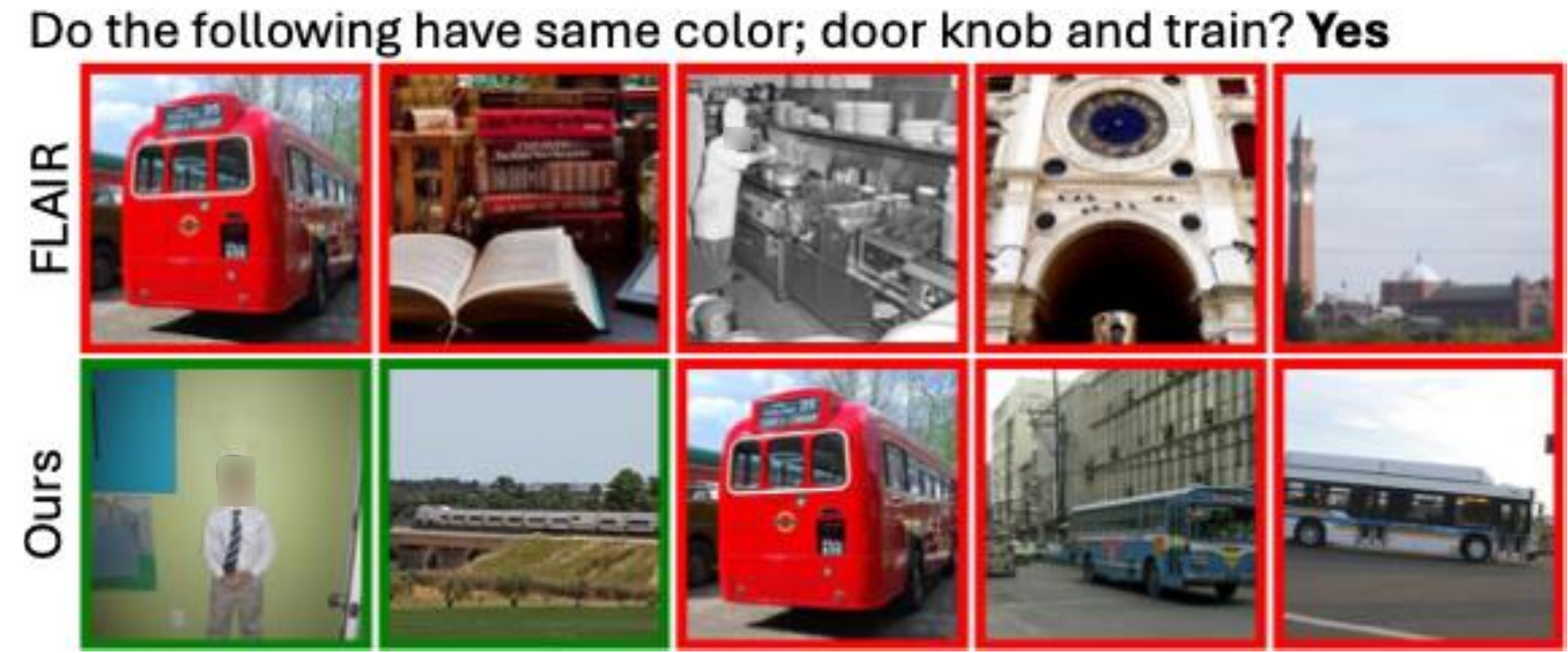}
    \caption{
        Example from RetVQA \cite{penamakuri2023answer}, where the task is to retrieve relevant images for a question that can be used as context for generative models. For each query, there are two relevant images and a set of distractors. Unlike FLAIR~\cite{xiao2025flair}, our model correctly retrieves images containing the door knob and the train that are relevant to the query.
    }
    \label{fig:retrieval-example}
\end{figure}

\begin{figure}[t]
    \centering
    \begin{minipage}{0.9\linewidth}
        \centering
        \includegraphics[width=\linewidth]{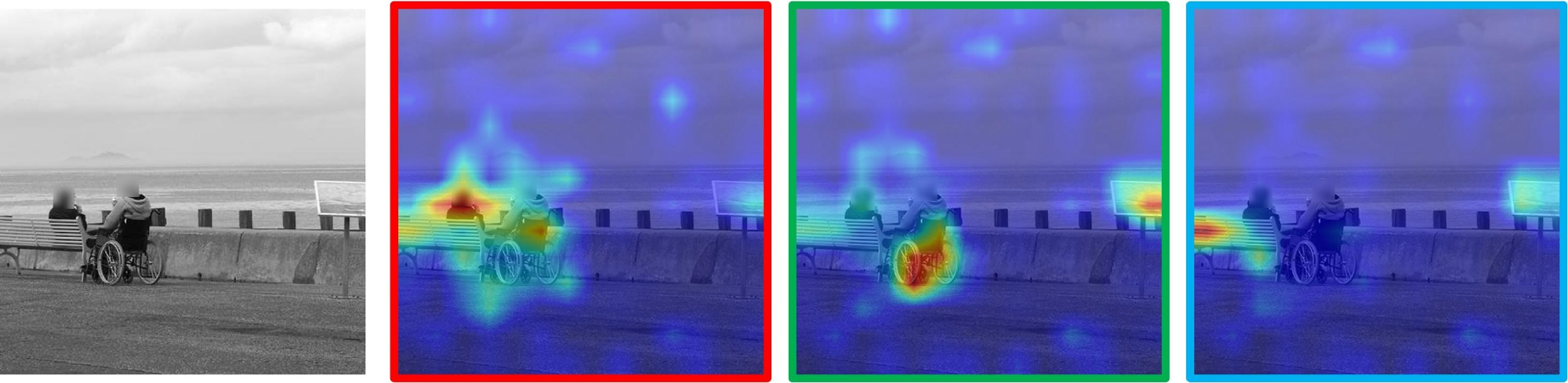}
        \subcaption{
            \textcolor[RGB]{255,0,0}{Two people} sitting, one on a \textcolor[RGB]{0,176,80}{wheelchair} and the other on a \textcolor[RGB]{0,176,240}{bench}.
        }
    \end{minipage}

    \begin{minipage}{0.9\linewidth}
        \centering
        \includegraphics[width=\linewidth]{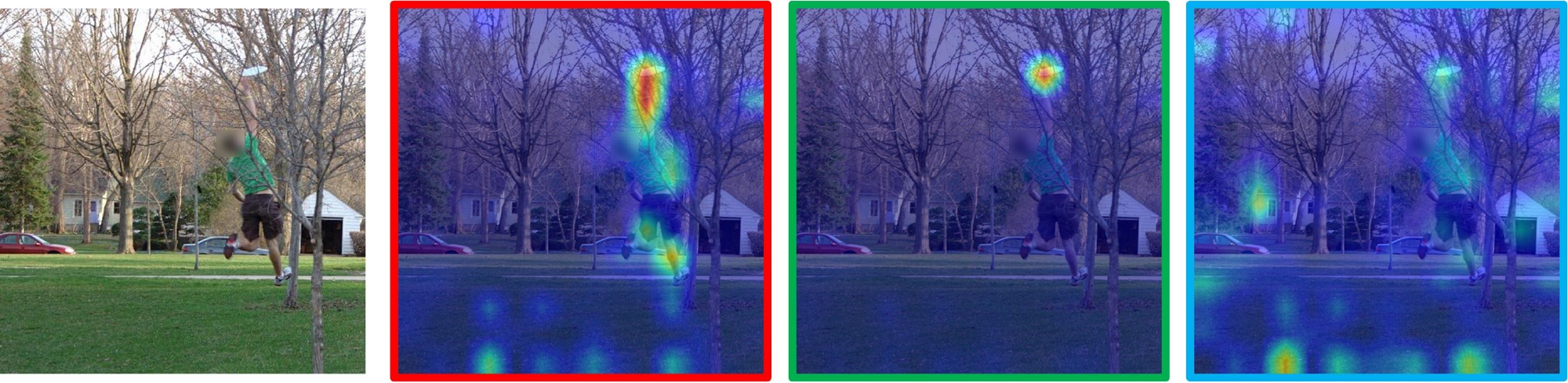}
        \subcaption{
            A \textcolor[RGB]{255,0,0}{young man jumping} to catch a \textcolor[RGB]{0,176,80}{frisbee} in the \textcolor[RGB]{0,176,240}{front yard}.
        }
    \end{minipage}
    \caption{
        Attention maps indicate good vision-text alignment.
    }
    \label{fig:qual_heatmap}
\end{figure}

\subsection{Training Data}
We train our model on the Open-30M dataset, which consists of CC3M \cite{sharma2018conceptual}, CC12M \cite{changpinyo2021conceptual}, and YFCC15M \cite{thomee2016yfcc100m}, similar to prior work~\cite{xiao2025flair, kim2025cosmos, bulat2024fff}. After accounting for broken URLs, the dataset contains 27 million images in total. We use the captions from \cite{zheng2024dreamlip}, which employed three VLMs to generate three short and three long captions for each image. 
To obtain auxiliary data for each image, we process a DreamLIP-generated~\cite{zheng2024dreamlip} caption by extracting object-centric phrases and using the OWL-V2~\cite{minderer2023scaling} model to detect bounding boxes corresponding to these phrases.

\subsection{Implementation}
\label{implementation}
We use standard CLIP vision and text encoders with embedding dimensions of $\mathrm{dim}_{t}=\mathrm{dim}_{v}=512$. 
Our vision encoder is based on ViT-B/16 and is trained with an input image resolution of $224\times224$. 
These parameters were applied across all experiments, including for comparison methods.

To compute the retrieval loss $\mathcal{L}_\mathrm{ret}$, we follow \cite{xiao2025flair} and construct positive text pairs by first breaking the six available captions into individual sentences and then randomly combining one to three of them to obtain a total of $K=8$ captions per image. We use a batch size of 6400.

For the multi-task decoder, the re-captioning objective is constructed by randomly sampling one of the three short full-length captions associated with an image. This strategy encourages the generation of coherent narratives. For other decoder objectives, such as grounded captioning and referring expression generation, we create localized training pairs by identifying an object-centric phrase and its corresponding sentence within one of the original captions. VQA pairs are generated for each image using the Phi-3.5-mini-instruct~\cite{abdin2024phi} model. For self-distillation, we generate $N_\mathrm{glob}=1$ global and $N_\mathrm{loc}=6$ local views of each image. Projection matrices $W^S$ and $W^T$ map the model's embeddings to a higher, 65,536-dimensional space before aligning the student and teacher representations via the self-distillation loss $\mathcal{L}_\mathrm{sd}$. 
The EMA momentum values are set to $m_\mathrm{sd}=0.996$ and $m_c=0.9$ for the teacher weights and centers updates, respectively. When combining all losses in our multi-task objective, the learnable task weighting parameters are all initialized to $\sigma_\mathrm{task} = 1$. We discard the teacher model and the multi-task decoder and report the performance using the vision and text encoder of the student model. Comprehensive implementation details, including all hyperparameters, are provided in the supplementary material.

\subsection{Zero-Shot Retrieval}
In Table~\hyperlink{tab:retrieval_full}{1a}, we compare the performance of our proposed approach against relevant methods on standard image-text retrieval benchmarks, including MSCOCO \cite{lin2014microsoft} and Flickr30k \cite{plummer2015flickr30k}, where each image has 5 global captions associated with it. We follow the evaluation setup in \cite{xiao2025flair}.
Our method demonstrates significant improvements over existing state-of-the-art models on text-to-image retrieval. On the MSCOCO benchmark (T2I, R@1), our approach outperforms FLAIR ~\cite{xiao2025flair} by 2.2pp and COSMOS \cite{kim2025cosmos} by 3pp. It also noteworthy that our method outperform SigLIP 2~\cite{tschannen2025siglip2} by 3pp and Perception Encoder~\cite{bolya2025perception} by 5.6pp, which are trained on 5-10 billion images with paired text.

\subsection{Fine-Grained and Long-Text Retrieval}
To analyze the fine-grained understanding capabilities of our model, we adopt the zero-shot retrieval tasks proposed by \cite{xiao2025flair} using DOCCI-FG~\cite{onoe2024docci} and IIW-FG~\cite{garg2024imageinwords}. These datasets are uniquely suited for this purpose, as they consist of densely annotated images where each caption is broken down into individual sentences. This compositional structure provides a challenging test, as it requires the model to comprehend specific, fine-grained details described in each sentence rather than general image-level descriptions found in MSCOCO and Flickr30k. Our model demonstrates superior fine-grained understanding by outperforming existing state-of-the-art methods on these benchmarks (see Table~\hyperlink{tab:retrieval_full}{1b}). Specifically, on the DOCCI-FG benchmark, we improve text-to-image R@1 by 2pp, and by 2.3pp on the IIW-FG benchmark compared to FLAIR~\cite{xiao2025flair}. This result strongly indicates that our diverse supervision framework, enables the model to learn more granular and detailed visual representations compared to other approaches. 
We also provide qualitative results in Figure \ref{fig:qual_heatmap}.
We also benchmark our model's performance on the challenging task of long text-to-image retrieval (Table~\hyperlink{tab:retrieval_full}{1c}). While prior state-of-the-art methods often extend the text encoder's sequence length to accommodate longer queries, we deliberately maintain the standard CLIP context length of 77 tokens to ensure a fair comparison with methods like FLAIR. Even with this constraint, our model demonstrates superior performance. On the DCI dataset, it outperforms FLAIR by 4.2pp on text-to-image retrieval (R@1) and improves image-to-text retrieval (R@1) by 3.8pp. These results indicate that our model is inherently more adept at matching long, detailed textual queries to their relevant images without requiring architectural modifications for extended text inputs.

\subsection{Zero-Shot Retrieval for VQA}
Traditionally, retrieval benchmarks have primarily been limited to textual queries that are descriptive captions, which fails to capture the full spectrum of real-world user interactions where queries are frequently phrased as questions in multi-modal RAG systems. State-of-the-art models trained on billion-scale datasets often exhibit robustness to such queries. We hypothesize that the sheer scale and noisy nature of their training datasets prevent the text encoder from overfitting to a rigid caption-like query structure. However, this is a common failure mode for models trained on smaller, curated re-captioned datasets.
To validate our method's robustness against this overfitting, we measure its performance on the Retrieval VQA task \cite{penamakuri2023answer} built on top of the Visual Genome dataset \cite{krishna2017visual}. The task is to retrieve relevant images that will provide correct context to answer a given question, as shown in Figure \ref{fig:retrieval-example}. We report R@2 numbers while showing our model's performance on different question types in Table \ref{tab:retrieval_vqa}. 
\begin{table}[t]
    \centering
    \scriptsize
    \caption{
         R@2 for zero-shot retrieval for VQA task on \cite{penamakuri2023answer} dataset build on top of VG \cite{krishna2017visual}. The best and second-best results \textbf{bolded} and \underline{underlined} respectively. 
    }
    \resizebox{\columnwidth}{!}{
        \begin{tabular}{lc|ccccc|c}
            \toprule
            \multirow{2}{*}{Method} & Data
            & \multicolumn{6}{c}{\textbf{T2I R2}} \\
            & size & Count & Color & Attribute & Shape & Relation & Average \\
            \toprule
            FLAIR~\cite{xiao2025flair} & 30M & 57.6 & 39.1 & 44.4 & 49.0 & 47.7 & 47.6\\
            COSMOS~\cite{zheng2024dreamlip} & 30M & 60.7 & 42.4 & 46.1 & 51.6 & 48.9 & 49.9\\
            \rowcolor{lightgold}
            GoldiCLIP & 30M & \textbf{66.8} & \underline{48.5} & \underline{51.5} & \underline{56.4} & {55.9} & \underline{55.8} \\
            \midrule
            OpenCLIP \cite{cherti2023reproducible} & 2B & 62.1 & 45.0 & 47.7 & 51.3 & 48.5 & 50.9 \\
            PE$_\textrm{core}$\cite{bolya2025perception} & 5B & 65.1 & 67.4 & 48.9 & 52.4 & \underline{57.5} & 52.9 \\
            SigLIP \cite{zhai2023sigmoid} & 10B & 64.1 & 46.5 & 51.1 & 53.2 & 53.5 & 53.5 \\
            SigLIP~2 \cite{tschannen2025siglip2} & 10B & 66.6 & \textbf{49.1} & \textbf{54.7} & \textbf{56.7} & \textbf{57.7} & \textbf{57.0} \\
            \bottomrule
        \end{tabular}
    }
    \label{tab:retrieval_vqa}
    \vspace{-1em}
\end{table}

Our method achieves an average performance improvement of 5pp over existing approaches trained on comparable data scales (30M), while also outperforming models trained on larger datasets (SigLIP~\cite{zhai2023sigmoid}, PE~\cite{bolya2025perception}) and is comparable to SigLIP 2~\cite{tschannen2025siglip2}. 
This improved generalization can be attributed to our enhanced multi-task decoder with VQA capabilities, which helps maintain robust performance across diverse query formats even when trained on caption-heavy datasets. By progressively bridging the gap to state-of-the-art models trained on large-scale datasets, our method demonstrates both superior performance and enhanced generalization capabilities.

\subsection{Zero-Shot Semantic Segmentation}

\begin{table}[tb]
    \centering\scriptsize
    \caption{
        Mean intersection over union (mIoU) for zero-shot semantic segmentation.
        The best and second-best results are \textbf{bolded} and \underline{underlined}.
        $^*$No attention sink.
    }
    \resizebox{\linewidth}{!}{
        \begin{tabular}{lc|ccccc|c}
            \toprule
            Method & 
            \makecell{Data\\size}
            & VOC20
            & Cityscapes
            & Context59
            & ADE20K
            & COCO-Stuff
            & Average\\
            \toprule
            SigLIP~\cite{zhai2023sigmoid} & 30M & 14.5 & 5.5 & 5.8 & 2.2 & 3.8 & 6.4 \\
            DreamLIP~\cite{zheng2024dreamlip} & 30M & 1.8 & 0.9 & 0.4 & 0.1 & 0.1 & 0.7 \\
            COSMOS~\cite{kim2025cosmos}  & 30M & 53.6 & 13.9 & 15.7 & 8.5 & 10.7 & 20.0 \\
            FLAIR~\cite{xiao2025flair}  & 30M & \underline{73.0} & 13.6 & 18.6 & 10.4 & 13.3 & 25.8 \\
            FLAIR$^*$~\cite{xiao2025flair} & 30M & {70.3} & {26.2} & {26.4} & {14.7} & {18.8} & {31.3} \\
            \rowcolor{lightgold}
            GoldiCLIP & 30M & 70.0 & \underline{28.9} & \underline{29.5} & {16.4} & {19.4} & {32.8} \\
            \midrule
            CLIP~\cite{radford2021learning} & 400M & 41.8 & 5.5 & 9.2 & 3.2 & 4.4 & 12.8 \\
            OpenCLIP~\cite{cherti2023reproducible} & 2B & 47.2 & 5.1 & 9.0 & 2.9 & 5.0 & 13.9 \\
            MetaCLIP~\cite{xiao2025flair} & 2.5B & 35.4 & 5.0 & 8.1 & 2.2 & 4.3 & 11.0 \\
            PE$_\textrm{core}$~\cite{bolya2025perception} & 5B & \textbf{74.3} & 24.0 & \textbf{30.1} & \underline{17.2} & \textbf{20.0} & \underline{33.1} \\
            SigLIP~\cite{zhai2023sigmoid} & 10B & 66.1 & \textbf{29.1} & 26.4 & 16.3 & 18.0 & 31.2 \\
            SigLIP 2~\cite{tschannen2025siglip2} & 10B & \underline{73.0} & {27.9} & {28.4} & \textbf{17.8} & \underline{19.9} & \textbf{33.4} \\
            \bottomrule
        \end{tabular}
    }
    \label{tab:zs_segmentation}
    \vspace{-2em}
\end{table}

In Table \ref{tab:zs_segmentation}, we show the performance of our model on the zero-shot segmentation task. Similar to prior benchmarks, we compare our model against other methods using similar data scales as well as state-of-the-art methods. To compute segmentation maps, we follow the approach used by \cite{xiao2025flair}. Similar to \cite{xiao2025flair}, during training we add the zero vector to the keys and values matrices in the cross-attention pooling, which we call attention sinks. This is primarily done to help with cross-modal alignment in a contrastive setting. When computing zero-shot segmentation, we do not use this attention sink. To ensure a fair comparison, we report the \cite{xiao2025flair} numbers without the attention sink. 
GoldiCLIP on average outperforms all models trained on comparable data, with a 1.5pp improvement over the previous best \cite{xiao2025flair}. Additionally, GoldiCLIP is highly competitive with state-of-the-art models trained on much larger datasets. This performance can be attributed to our multi-task supervision framework that uses a text-conditioned self-distillation approach, which enables learning more robust and generalizable representations. 

\subsection{Zero-Shot Classification}
\begin{table*}[!t]
    \centering
    \scriptsize
    \caption{Top-1 accuracy for zero-shot classification. 
        The best and second-best results per section are \textbf{bolded} and \underline{underlined}.
    }
        \begin{tabular}{lc|cccccccccc|c}
            Method 
            & \makecell{Data\\size}
            & \rotatebox[origin=lb]{90}{{Food-101}} 
            & \rotatebox[origin=lb]{90}{{CIFAR-10}} 
            & \rotatebox[origin=lb]{90}{{CIFAR-100}} 
            & \rotatebox[origin=lb]{90}{{SUN397}} 
            & \rotatebox[origin=lb]{90}{{Cars}} 
            & \rotatebox[origin=lb]{90}{{Aircraft}} 
            & \rotatebox[origin=lb]{90}{{DTD}} 
            & \rotatebox[origin=lb]{90}{{Pets}} 
            & \rotatebox[origin=lb]{90}{{Flowers}} 
            & \rotatebox[origin=lb]{90}{{ImageNet}} 
            & \rotatebox[origin=lb]{90}{{Average}} \\
            \toprule
            CLIP~\cite{radford2021learning} & 30M & 61.3 & 92.2 & 66.9 & 62.2 & 19.3 & 5.7 & 30.9 & 49.3 & 43.4 & 50.0 & 48.1 \\
            SigLIP~\cite{zhai2023sigmoid} & 30M & 64.2 & 91.0 & 67.6 & {64.0} & 22.0 & 5.7 & 33.5 & 53.3 & 43.6 & 51.0 & 49.6 \\
            DreamLIP~\cite{zheng2024dreamlip} & 30M & \textbf{75.4} & {92.3} & \underline{70.7} & 63.7 & {22.7} & \textbf{7.9} & \underline{33.9} & \textbf{64.1} & {51.1} & \textbf{58.1} & {54.0} \\
            FLAIR~\cite{xiao2025flair} & 30M & {72.5} & \underline{93.1} & {69.6} & {66.9} & {31.1} & \underline{7.2} & \textbf{37.3} & {55.6} & {48.4} & {56.6} & {53.8} \\
            COSMOS~\cite{kim2025cosmos} & 30M & 73.9 & \textbf{93.4} & \textbf{72.3} & \textbf{67.8} & \underline{31.2} & 5.9 & \textbf{37.3} & \underline{62.9} & \textbf{54.6} & \underline{57.6} & \textbf{55.7} \\
            \rowcolor{lightgold}
            GoldiCLIP & 30M & \underline{74.0} & 92.9 & \underline{70.7} & \underline{67.2} & \textbf{34.5} & {6.6} & \underline{33.9} & 62.7 & \underline{51.4} & 57.4 & \underline{55.1} \\
            \midrule
            OpenCLIP~\cite{cherti2023reproducible} & 2B & {86.2} & 94.8 & 76.5 & {70.0} & 87.4 & 25.8 & 54.9 & 89.5 & 69.8 & 70.2 & 72.5 \\
            MetaCLIP~\cite{xu2023demystifying} & 2.5B & 88.3 & {95.7} & {79.0} & 68.5 & {82.9} & {30.3} & {62.1} & {91.7} & {73.9} & {72.1} & {74.5} \\
            Llip~\cite{lavoie2024modeling} & 2.5B & {89.0} & \underline{95.7} & \underline{81.4} & {70.9} & {88.2} & {41.5} & {63.7} & {93.5} & {74.9} & {75.3} & {77.4} \\
            PE$_\textrm{core}$~\cite{bolya2025perception} & 5B & \underline{92.5} & \textbf{97.0} & \textbf{83.1} & \textbf{74.0} & \underline{92.1} & \textbf{57.0} & \underline{66.4} & \underline{94.6} & \textbf{86.5} & \textbf{78.4} & \textbf{82.2} \\
            SigLIP~\cite{zhai2023sigmoid} & 10B & 91.6 & 92.3 & 72.2 & 70.0 & 90.8 & 44.0 & 64.7 & 94.2 & 85.2 & 76.2 & 78.1 \\
            SigLIP2~\cite{tschannen2025siglip2} & 10B & \textbf{92.8} & 94.8 & 76.9 & \underline{72.7} & \textbf{93.4} & \underline{54.8} & \textbf{67.2} & \textbf{95.4} & \underline{85.7} & \underline{78.2} & \underline{81.2} \\
            \bottomrule
        \end{tabular}
    \label{tab:zs_classification}
\end{table*}

We present the results of zero-shot classification on 10 standard classification datasets in Table \ref{tab:zs_classification}. While our performance is comparable to methods trained on similar data scales, there remains a significant performance gap when compared to state-of-the-art models such as SigLIP 2 \cite{tschannen2025siglip2} and PE \cite{bolya2025perception}. We hypothesize that this gap arises primarily due to two factors. First, state-of-the-art models are trained on significantly larger datasets (5-10 billion image samples), which encompass a broader range of diverse concepts compared to the 30 million image samples in our dataset. Second, the re-captions generated using Vision-Language Models (VLMs) or Large Language Models (LLMs) tend to be more coarse-grained. This lack of fine-grained concepts in the re-captioned dataset limits its ability to generalize effectively on the zero-shot classification task. These factors collectively contribute to the observed performance disparity.

\subsection{Ablation Study}

In Table \ref{tab:main_ablation}, we present a performance ablation study of our model across various components, with all models trained on the CC3M-recap dataset. The cross-attention pooling (row 1) aligns with previous work \cite{xiao2025flair}. Notably, the addition of individual components provides complementary benefits. Specifically, enhancing the multi-task decoder with VQA improves performance on question-based retrieval tasks and helps avoid overfitting to caption-like queries when trained on larger datasets. Similarly, the model's segmentation performance enhances spatial understanding, as evidenced by the addition of self-distillation, which improves performance by 2.9pp. Furthermore, integrating all supervision components into a joint framework boosts the model's performance by 7.4pp for MS-COCO text-to-image retrieval (R@1) and 3.6pp for fine-grained retrieval (T2I, R@1), among other tasks. To isolate the effect of our text-conditioned self-distillation, we apply it to FLAIR as standalone addition and observe improvements of 4.9 and 4.3 pp on MSCOCO T2I and I2T retrieval respectively (Table~\ref{tab:prior_methods}), confirming that the gains stem from the proposed distillation mechanism rather than from the combination of other components. We discuss additional details, including the experimental settings and design choices related to this ablation study in the supplementary material. 
\begin{table}[t] %
    \centering
    \scriptsize
    \caption{
        Ablation study on different components of our method on the CC3M-recap dataset. \textbf{CA}: Cross Attention Pooling, \textbf{DC}: Multi-Task Decoder without VQA objective, \textbf{VD}: Multi-Task Decoder with VQA, \textbf{UN}: Uncertainty-based Task Weighting, \textbf{SD}: Text-Conditioned Self-Distillation.
    }
    \label{tab:main_ablation}
    \begin{tabular}{l|ccccccc} %
        \toprule
        \multirow{2}{*}{Method} & \multicolumn{2}{c}{COCO} & \multicolumn{2}{c}{DOCCI} & RVQA & VOC20 & ImageNet \\
        & T2I & I2T & T2I & I2T & T2I & mIOU & Top-1 \\
        \midrule
        \phantom{$+$}CA & 37.7 & 51.6 & 15.1 & 35.7 & 51.0 & 59.0 & 36.2 \\
        $+$DC & 42.1 & 56.4 & 16.7 & 40.3 & 51.4 & 50.6 & 37.5 \\
        $+$VD & 42.1 & 56.9 & 16.7 & 39.0 & 51.7 & 53.9 & 37.6 \\
        $+$UN & 42.9 & 58.2 & 17.6 & 42.3 & 52.0 & 55.1 & 38.8 \\
        $+$SD & 45.1 & 58.6 & 18.7 & 43.0 & 51.7 & 62.9 & 40.1\\
        \bottomrule
    \end{tabular}
\end{table}

\begin{table}[t] %
    \centering
    \scriptsize %
    \caption{Comparison of Prior Self-distillation methods to our text conditioned self-distillation applied to FLAIR. All methods trained on CC3M-recap dataset. Recall @1 reported for both I2T and T2I. Best results bolded.
    } %
    \label{tab:prior_methods}
    \begin{tabular}{l|cc|cc}
        \toprule
        \multirow{2}{*}{Method} & \multicolumn{2}{c}{COCO} & \multicolumn{2}{c}{Flickr30k} \\
            & T2I & I2T & T2I & I2T \\
        \midrule
        SILC & 48.6 & 35.4 & 79.0 & 62.1 \\
        COSMOS & 53.1 & 40.1 & 84.1 & 68.6 \\
        FLAIR & 51.6 & 37.7 & 78.7 & 65.5 \\
        \textbf{FLAIR + TC-SD} & \textbf{56.5} & \textbf{42.0} & \textbf{87.5} & \textbf{70.3}\\
        \bottomrule
    \end{tabular}
\end{table}

\section{Conclusion}
\label{sec:conclusion}
GoldiCLIP achieves state-of-the-art performance among methods trained on comparable data scales across retrieval, fine-grained retrieval, question-based retrieval, and semantic segmentation benchmarks. Among our contributions, text-conditioned self-distillation stands out as particularly impactful: by extending local-to-global consistency learning to text-conditioned features, it directly improves the representations that matter most for fine-grained retrieval, yielding substantial gains even when applied to existing frameworks. Routing decoder tasks through the text encoder, distilling both text-conditioned and text-agnostic features, and automatically balancing losses across scales are individually small design decisions, but together they account for the consistent improvements we observe across retrieval, fine-grained understanding, and dense prediction tasks.
Our framework has clear limitations. Zero-shot classification remains constrained by the limited concept diversity in smaller datasets, a gap that richer supervision alone does not close. Additionally, our experiments are limited to a single ViT-B/16 backbone; understanding how these supervision interactions behave at larger model scales is an open question. Our work suggests that how supervision signals are adapted and integrated is a complementary research direction to the design of the signals themselves, and we hope that our findings encourage further research in this direction.

\clearpage
\bibliographystyle{splncs04}
\bibliography{main}

\clearpage
\setcounter{page}{1}
\maketitlesupplementary

\section{Training Data}

\subsection{Images}
Our training images are sourced from CC3M \cite{sharma2018conceptual}, CC12M \cite{changpinyo2021conceptual}, and YFCC15M \cite{thomee2016yfcc100m}. We follow prior work and use these images directly without any additional filtering or modification.

\subsection{Captions}
To construct a comprehensive and diverse caption set, we incorporate the raw captions as well as synthetic captions provided by DreamLIP~\cite{zheng2024dreamlip}. Specifically, for each image we include one raw caption, three synthetic short captions, and three synthetic long captions.

DreamLIP generates these captions using a combination of InstructBLIP, ShareGPT4V, and LLaVA-1.5.
Their pipeline uses the following prompts:
\begin{itemize}
    \item \textit{``Describe the image in details''} for long captions
    \item \textit{``Describe the image in short:''} for short captions.
\end{itemize}
We refer readers to the DreamLIP paper for further details on their caption generation procedure.

\subsection{Auxiliary Data}
To enrich visual supervision, we supplement the dataset with object bounding boxes, region-level caption phrases and synthetic visual questions and answers.

\textbf{Object Bounding Boxes.}
We leverage OWLv2~\cite{minderer2023scaling}, an open-vocabulary detector, to predict bounding boxes. Following prior work \cite{minderer2023scaling, tschannen2025siglip2}, we first extract region-level phrases using named entity recognition (NER) applied to the long captions generated by LLaVA-1.5. These extracted region phrases serve as text queries for OWLv2. For each query, OWLv2 returns bounding boxes, confidence scores, and region captions.

\textbf{VQA Generation.}
We generate additional VQA supervision using Phi-3.5-mini-instruct~\cite{abdin2024phi} with the following prompt template:
\begin{quote}
\small
Given the following image caption: \\
``\texttt{ref\_caption}'' \\
Please generate five short questions and corresponding short answers that someone might ask about this image. The questions should be relevant to the content of the caption, and the answers be very short, concise, and factual based on the caption. Only generate questions and answers that are strictly answerable using the provided caption. Do not invent details or use outside knowledge.
\end{quote}

\sloppy
Here, \texttt{ref\_caption} is replaced with the short caption produced by ShareGPT4V. We set the maximum generation length to 500 tokens. This results in approximately 3--5 question--answer pairs per image. The creation of this additional VQA data is in the same spirit as the VLM based re-captioning used by prior methods and is considered as text augmentations, as no additional images are used.

\section{Implementation in Detail}
\label{sec:implementation_details}

\begin{table}[t]
    \centering\scriptsize
    \caption{
        Pre-training hyper-parameters for our method. 
    }
        \begin{tabular}{lcc}
            \toprule
            Config & CC3M-recap & Open-30M \\
            \midrule 
            Batch size & $2,560$ & $6,400$ \\
            Optimizer & \multicolumn{2}{c}{AdamW}  \\
            Learning rate & \multicolumn{2}{c}{$5\times10^{-4}$} \\
            Weight decay & $0.5$ & $0.2$ \\
            Adam $\beta$ & \multicolumn{2}{c}{$\beta_1=0.9, \beta_2=0.98$} \\
            Adam $\epsilon$ & $1\times10^{-8}$ & $1\times10^{-6}$ \\
            Total epochs & \multicolumn{2}{c}{$35$} \\
            Warm up & \multicolumn{2}{c}{$2,000 \text{ (steps)}$} \\
            \makecell{Learning rate scheduler} & \multicolumn{2}{c}{cosine decay}  \\
            Teacher momentum $m_\textrm{sd}$ & \multicolumn{2}{c}{0.996} \\
            Center momentum $m_\textrm{c}$ & \multicolumn{2}{c}{0.9} \\
            \bottomrule
        \end{tabular}
    \label{tab:hyperparam}
\end{table}

As mentioned in Section~\ref{implementation}, we use standard CLIP vision and text encoders. Text encoder has embedding dimensions of $\mathrm{dim}_{t}=512$. 
Our vision encoder is based on ViT-B/16 and is trained with an input image resolution of $224\times224$. 
These parameters were applied across all experiments, including for comparison methods. The hidden dimension of vision encoder is 768 (as in ViT-B/16). Similar to \cite{xiao2025flair} we project this to $\mathrm{dim}_{v}=512$. Additionally in order to perform the cross attention pooling we use a secondary projection, from the the vision encoder. This projected visual tokens are used as the key and value tokens, in creating the text conditioned representations. The multi-modal decoder used for training has hidden dimension of $\mathrm{dim}_{dec}=512$. This is done so that the vision tokens after projection (corresponding to key projection mentioned above) can flow into the decoder without the need for additional projection. The decoder consists of 8 layers with 8 heads, with the same vocabulary as the text encoder. 

To compute the retrieval loss $\mathcal{L}_\mathrm{ret}$, we follow \cite{xiao2025flair} and construct positive text pairs by first breaking the six available captions into individual sentences and then randomly combining one to three of them to obtain a total of $K=8$ captions per image. We use a batch size of 6,400 when training on Open-30M and 2,560 for training on CC3M-recap. When creating text conditioned representation in a batch, it becomes computationally challenging to create text-conditioned representation on all the negative images for each image in a batch. We follow the techniques used by \cite{xiao2025flair} to downsample the number of negatives in a batch, to only use one of negatives from the set of $K$ captions associated with the negative image. We report all the other parameters related to training the model for the main experiments as well as the ablations in Table~\ref{tab:hyperparam}. 
In the final experiments, we found that weighting caption loss with a value of 2 further boosts performance. This indicates that while task balancing provides significant benefits by largely mitigating the need for extensive hyperparameter searches, fine-tuning specific hyperparameters can still offer incremental improvements. All our experiments are run on MI300X AMD gpu's. 

\section{Task Balancing}
\begin{figure}[h!]
    \centering
    \includegraphics[width=0.8\linewidth]{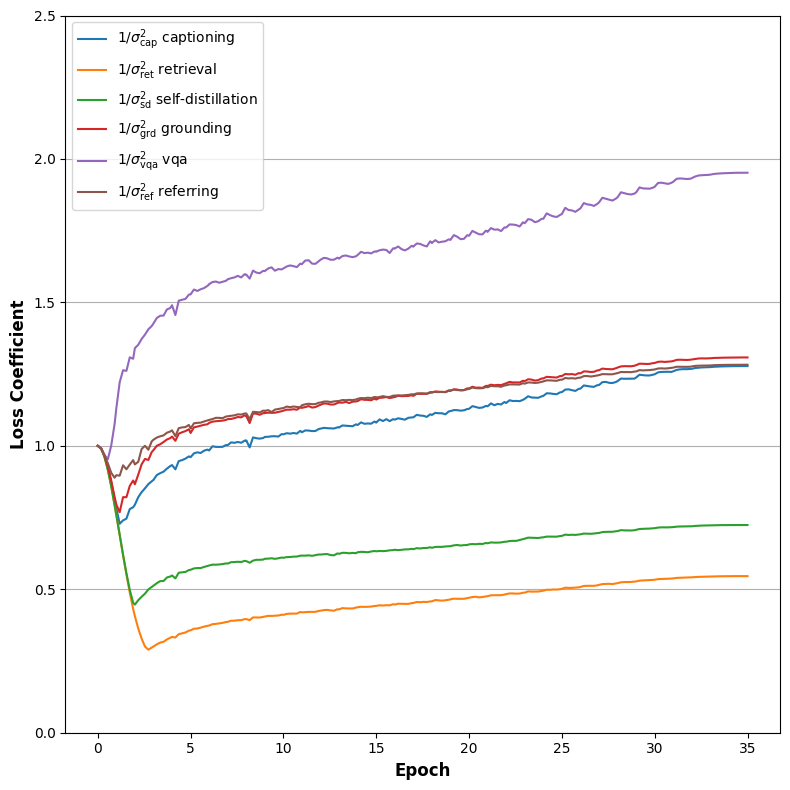}
    \caption{
        Values of task coefficients changing over training epochs.
    }
    \label{fig:task-balancing}
\end{figure}

Figure~\ref{fig:task-balancing} demonstrates how the learnable task-balancing coefficients $1/\sigma_\textrm{task}^2$ change throughout training. In order to analyse the effect of task-balancing on final performance, we compare it with the model trained with uniform loss weights (while supervision remains the same). When trained on CC3M-recap dataset, we observe that with uniform weight the R@1 for text-to-image retrieval on MSCOCO dataset is 44.1 whereas with uncertainty task weighting it is 45.0

\section{Text-agnostic Performance}

\begin{table*}[!t]
    \hypertarget{tab:retrieval_full}{} 
    \caption{
        Our model remains competetive on various retrieval benchmarks even when evaluated in text-agnostic mode, denoted as GoldiCLIP$_\mathrm{ta}$.
    }
    \resizebox{\textwidth}{!}{
        \begin{tabular}{lc|cccccccc|cccc|cccc}
            
            & & \multicolumn{8}{|c|}{(a) Standard Benchmarks} & \multicolumn{4}{|c|}{(b) Fine-Grained} & \multicolumn{4}{|c}{(c) Long Text} \\
            \toprule
            \multirow{3}{*}{Method} & \multirow{3}{*}{\makecell{Data\\size}} & \multicolumn{4}{c}{MSCOCO} & \multicolumn{4}{c|}{Flickr30k} & \multicolumn{2}{c}{DOCCI-FG} & \multicolumn{2}{c|}{IIW-FG} & \multicolumn{2}{c}{DCI} & \multicolumn{2}{c}{Urban 1k} \\
            \cmidrule(lr){3-6}
            \cmidrule(lr){7-10}
            \cmidrule(lr){11-12}
            \cmidrule(lr){13-14}
            \cmidrule(lr){15-16}
            \cmidrule(lr){17-18}
            & & \multicolumn{2}{c}{T2I} & \multicolumn{2}{c}{I2T} & \multicolumn{2}{c}{T2I} & \multicolumn{2}{c|}{I2T} & T2I & I2T & T2I & I2T & T2I & I2T & T2I & I2T \\
            & & R1 & R5 & R1 & R5 & R1 & R5 & R1 & R5 & R1 & R1 & R1 & R1 & R1 & R1 & R1 & R1 \\
            \toprule
            DreamLIP~\cite{zheng2024dreamlip} & 3M & 29.8 & 55.4 & 40.8 & 68.4 & 53.6 & 78.4 & 69.2 & 91.5 & 10.3 & 23.3 & 22.7 & 59.2 & - & - & - & - \\
            FLAIR~\cite{xiao2025flair} & 3M & 37.7 & 64.5 & 51.6 & 77.2 & 65.7 & 86.8 & 78.7 & 95.2 & 15.2 & 35.7 & 30.5 & 70.6 & 50.3 & 47.3 & 69.5 & 63.5\\ 
            COSMOS~\cite{kim2025cosmos} & 3M & 40.1 & 66.8 & 53.1 & 78.3 & 68.6 & 88.7 & 84.1 & 96.5 & 14.2 & 32.9 & 28.9 & 70.9 & 52.5 & 50.2 & 73.2 & 67.1 \\ 
            \rowcolor{lightgold}
            GoldiCLIP & 3M & {45.1} & {71.2} & {58.6} & {82.4} & {73.8} & {91.2} & {88.7} & {97.4} & {18.7} & {43.0} & {35.4} & {79.4} & 57.7 & 52.8 & 79.3 & 72.3 \\ 
            \midrule
            SigLIP~\cite{zhai2023sigmoid} & 30M & 49.0 & 74.5 & 64.1 & 86.2 & 76.7 & 93.2 & 90.5 & 98.3 & - & - & - & - & - & - & - & - \\
            DreamLIP ~\cite{zheng2024dreamlip} & 30M & 44.8 & 69.8 & 62.3 & 84.5 & 73.3 & 91.8 & 89.9 & 99.0 & 21.6 & 51.2 & 37.5 & 85.3 & - & - & - & - \\
            COSMOS ~\cite{kim2025cosmos} & 30M & 52.5 & 77.2 & 68.0 & 87.8 & 80.3 & 95.3 & 92.9 & \textbf{99.4} & 23.1 & 57.3 & 40.1 & 90.2 & 67.7 & 63.7 & 88.0 & 82.7 \\
            FLAIR ~\cite{xiao2025flair} & 30M & 53.3 & 77.5 & 68.0 & 87.8 & 81.1 & 94.9 & 94.7 & 99.3 & 25.0 & 59.0 & 41.7 & 91.5 & 66.3 & 61.3 & 87.7 & \textbf{83.6}\\
            \rowcolor{lightgold}
            GoldiCLIP & 30M & \textbf{55.5} & \textbf{78.9} & \textbf{70.3} & \textbf{89.3} & \textbf{83.0} &\textbf{95.9} & \textbf{94.8} & 99.1 & \textbf{27.0} & \textbf{60.2} & \textbf{44.0} & \textbf{90.9} & \textbf{71.9} & \textbf{67.5} & \textbf{88.6} & 83.2 \\
            \rowcolor{lightgold}
            GoldiCLIP$_\mathrm{ta}$ & 30M & 52.3 & 77.2 &  67.7 & 88.0 & 80.3 & 95.0 & 93.2 & 99.2 & 22.9 & 55.0 & 38.5 & 89.2 & 69.2 & 65.9 & 85.8 & 83.0 \\
            \midrule
            LiT~\cite{zhai2022lit} & 100M &  32.7 & 57.8 & 51.9 & 75.7 & 60.9 & 84.6 & 80.0 & 95.9 & - & - & - & - & 40.9 & 41.7 & - & - \\
            LoTLIP~\cite{wu2024lotlip} & 100M & 38.1 & 63.8 & 59.7 & 81.5 &  65.2 & 88.0 & 86.9 & 97.8 & - & - & - & - & 61.0 & 62.1 & - & - \\
            Long-CLIP~\cite{zhang2024long} & 400M & 40.3 & 65.9 & 57.3 & 80.8 & 70.7 & 90.6 & 85.9 & 98.5 & - & - & - & - & 44.1 & 47.4 & 79.5 & 78.9 \\
            OpenCLIP~\cite{cherti2023reproducible} & 2B & 41.7 & 67.1 & 59.3 & 82.4 & 71.9 & 90.4 & 87.5 & 97.7 & 17.4 & 49.7 & 30.6 & 84.1 & 55.4 & 56.0 & 65.8 & 69.5 \\
            PE$_\textrm{core}$~\cite{bolya2025perception} & 5B & 49.9 & 74.8 & \textbf{70.3} & 88.9 & 81.3 & 95.4 & 93.5 & 99.0 & 22.4 & 61.8 & 36.8 & 90.7 & 60.0 & 60.2 & 56.7 & 60.3 \\
            SigLIP~\cite{zhai2023sigmoid} & 10B & 47.2 & 72.1 & 65.5 & 86.2 & 75.6 & 92.8 & 89.1 & 98.6 & 20.6 & 57.5 & 33.8 & 83.7 & 56.2 & 57.7 & 62.1 & 62.7 \\
            SigLIP 2~\cite{tschannen2025siglip2} & 10B & 52.5 & 76.6 & 70.0 & 88.3 & 80.0 & 94.5 & 91.8 & 98.8 & 23.2 & 61.2 & 37.0 & 91.3 & 55.7 & 55.7 & 57.8 & 60.3 \\
            \bottomrule
        \end{tabular}
     }
    \label{tab:retrieval_full_single}
\end{table*}

\begin{table*}[t]
    \centering\scriptsize
    \caption{
         Our model remains competetive on the retrieval for VQA task even when evaluated in text-agnostic mode, denoted as GoldiCLIP$_\mathrm{ta}$.
    }
        \begin{tabular}{lc|ccccc|c}
            \toprule
            \multirow{2}{*}{Method} & Data
            & \multicolumn{6}{c}{\textbf{T2I R2}} \\
            & size & Count & Color & Attribute & Shape & Relation & Average \\
            \toprule
            FLAIR~\cite{xiao2025flair} & 30M & 57.6 & 39.1 & 44.4 & 49.0 & 47.7 & 47.6\\
            COSMOS~\cite{zheng2024dreamlip} & 30M & 60.7 & 42.4 & 46.1 & 51.6 & 48.9 & 49.9\\
            \rowcolor{lightgold}
            GoldiCLIP & 30M & \textbf{66.8} & \underline{48.5} & \underline{51.5} & \underline{56.4} & {55.9} & \underline{55.8} \\
            \rowcolor{lightgold}
            GoldiCLIP$_\mathrm{ta}$  & 30M & 65.5 & 46.9 & 48.8 & 55.0 & 54.5 & 54.1 \\
            \midrule
            OpenCLIP \cite{cherti2023reproducible} & 2B & 62.1 & 45.0 & 47.7 & 51.3 & 48.5 & 50.9 \\
            PE$_\textrm{core}$\cite{bolya2025perception} & 5B & 65.1 & 67.4 & 48.9 & 52.4 & \underline{57.5} & 52.9 \\
            SigLIP \cite{zhai2023sigmoid} & 10B & 64.1 & 46.5 & 51.1 & 53.2 & 53.5 & 53.5 \\
            SigLIP~2 \cite{tschannen2025siglip2} & 10B & 66.6 & \textbf{49.1} & \textbf{54.7} & \textbf{56.7} & \textbf{57.7} & \textbf{57.0} \\
            \bottomrule
        \end{tabular}
    \label{tab:retrieval_vqa_single}
\end{table*}

In Tables~\ref{tab:retrieval_full_single} and \ref{tab:retrieval_vqa_single} we report the performance of our model in text-agnostic mode. In this mode, instead of using the text-condition image representation $\mathbf{v}_\mathrm{tc}$, we use the text-agnostic $\mathbf{v}_\mathrm{ta}=\mathbf{v}_\mathrm{cls}$. This mode may come useful when storing all image patch embeddings $\mathbf{v}_1, \ldots, \mathbf{v}_N$ is not feasible. While text-conditioning step (cross-attention to the text) is critical for getting the best possible performance using our model, GoldiCLIP remains competitive in the text-agnostic mode.

\section{Method pseudocode}

We provide PyTorch-like pseudocode describing the computation of supervision objectives in Listing~\ref{lst:pseudocode} to offer a more detailed overview of our method.

\begin{figure*}
    \begin{lstlisting}[language=Python, caption={GoldiCLIP pseudocode}, label={lst:pseudocode}]
    Inputs: batch of (images BxHxWx3, captions BxK, aux_data Bx...), learnable params, momentums m_c, m_sd
    
    # Contrastive objectives
    image_token, patch_tokens = params.student.V(global_views(images))
    image_key, image_value = params.student.project_key(patch_tokens), params.student.project_value(patch_tokens)
    text_tokens = params.student.T(captions)
    v_ta = params.student.project_key(image_token)
    
    L_ta, L_tc = 0, 0
    for i in range(B):
        for j in range(B):
            for k in range(K):
                ta_similarity = cosine(v_ta[i], text_tokens[j, k, -1])
                ta_similarity = - params.student.t * ta_similarity + params.student.b
                L_ta += 1 / (1 + exp(y(i, j, k) * ta_similarity))
    
                v_tc[i, j, k] = params.student.cross_attend(
                                    text_tokens[j, k, -1], 
                                    image_key[i],
                                    image_value[i]
                                )
                tc_similarity = cosine(v_tc[i, j, k], text_tokens[j, k, -1])
                tc_similarity = - params.student.t * tc_similarity + params.student.b
                L_tc += 1 / (1 + exp(y(i, j, k) * tc_similarity))
    L_ret = L_ta + L_tc
    
    # Decoder Objectives
    prefix[task] = {
        'caption': ...,
        'grounding': ...
        'referring': ...
        'vqa': ...
    }
    for task in tasks:
        task_tokens = params.student.T(prefix[task] + aux_data[task])
        preds = params.D(task_tokens, image_key)
        L_dec[task] = NLL(preds, task_tokens)
    \end{lstlisting}
\end{figure*}

\begin{figure*}
\begin{lstlisting}[language=Python, firstnumber=39]
# ...continued...
# Self-Distillation Objective
teacher = m_sd * params.student + (1 - m_sd) * teacher

student_image_token, student_patch_tokens = params.student.V(local_views(images))
teacher_image_token, teacher_patch_tokens = teacher.V(global_views(images))

student_image_key = params.student.project_key(student_patch_tokens)
student_image_value = params.student.project_value(student_patch_tokens)
teacher_image_key = params.teacher.project_key(teacher_patch_tokens)
teacher_image_value = params.teacher.project_value(teacher_patch_tokens)

student_vta = params.student.project_key(student_image_token)
teacher_vta = params.teacher.project_key(teacher_image_token)

for i in range(B):
    for j in range(B):
        for k in range(K):
            student_vtc[i, j, k] = params.student.cross_attend(
                                        text_tokens[j, k, -1],
                                        student_image_key[i], 
                                        student_image_value[i]
                                    )
            teacher_vtc[i, j, k] = teacher.cross_attend(
                                        text_tokens[j, k, -1],
                                        teacher_image_key[i], 
                                        teacher_image_value[i]
                                    )

# Project all features
student_vta = params.student.project(student_vta)
student_vtc = params.student.project(student_vtc)
teacher_vta = teacher.project(teacher_vta)
teacher_vtc = teacher.project(teacher_vtc)

# Average TC features over captions K'
student_logits_vtc = student_vtc.mean(over='captions')
teacher_logits_vtc = teacher_vtc.mean(over='captions')

# Center teacher features
teacher_logits_vta = teacher_vta - params.c_ta
teacher_logits_vtc = teacher_logits_vtc - params.c_tc

# Scale features and compute probabilities
student_logits_vta /= params.student_tau
student_logits_vtc /= params.student_tau
teacher_logits_vta /= params.teacher_tau
teacher_logits_vtc /= params.teacher_tau
student_probs_vta = softmax(student_logits_vta)
student_probs_vtc = softmax(student_logits_vtc)
teacher_probs_vta = softmax(teacher_logits_vta)
teacher_probs_vtc = softmax(teacher_logits_vtc)

L_sd = 0
for i in range(N_glob):
    for j in range(N_loc):
        L_sd += cross_entropy(teacher_probs_vta[i], student_probs_vta[j])
        L_sd += cross_entropy(teacher_probs_vtc[i], student_probs_vtc[j])

# Update centers
params.c_ta = m_c * params.c_ta + (1 - m_c) * teacher_vta.mean(over=['batch'])
params.c_tc = m_c * params.c_tc + (1 - m_c) * teacher_vtc.mean(over=['batch', 'captions'])

# Task balancing
L = 1 / (params.sigma_ret ** 2) * (L_ta + L_tc) + params.sigma_ret ** 2
  + 1 / (params.sigma_cap ** 2) * L_dec['captioning'] + params.sigma_cap ** 2
  + 1 / (params.sigma_grd ** 2) * L_dec['grounding'] + params.sigma_grd ** 2
  + 1 / (params.sigma_ref ** 2) * L_dec['referring'] + params.sigma_ref ** 2
  + 1 / (params.sigma_vqa ** 2) * L_dec['vqa'] + params.sigma_vqa ** 2
  + 1 / (params.sigma_sd ** 2) * L_sd + params.sigma_sd ** 2
\end{lstlisting}
\end{figure*}

\clearpage
\section{Improved alignment}
The figure below shows the attention visualization of GoldiCLIP model before (top row) and after applying our text-conditioned SD (bottom row), showing improved alignment behavior.

\begin{figure}[ht]
    \centering
    \begin{minipage}{\linewidth}
        \centering
        \includegraphics[width=\linewidth]{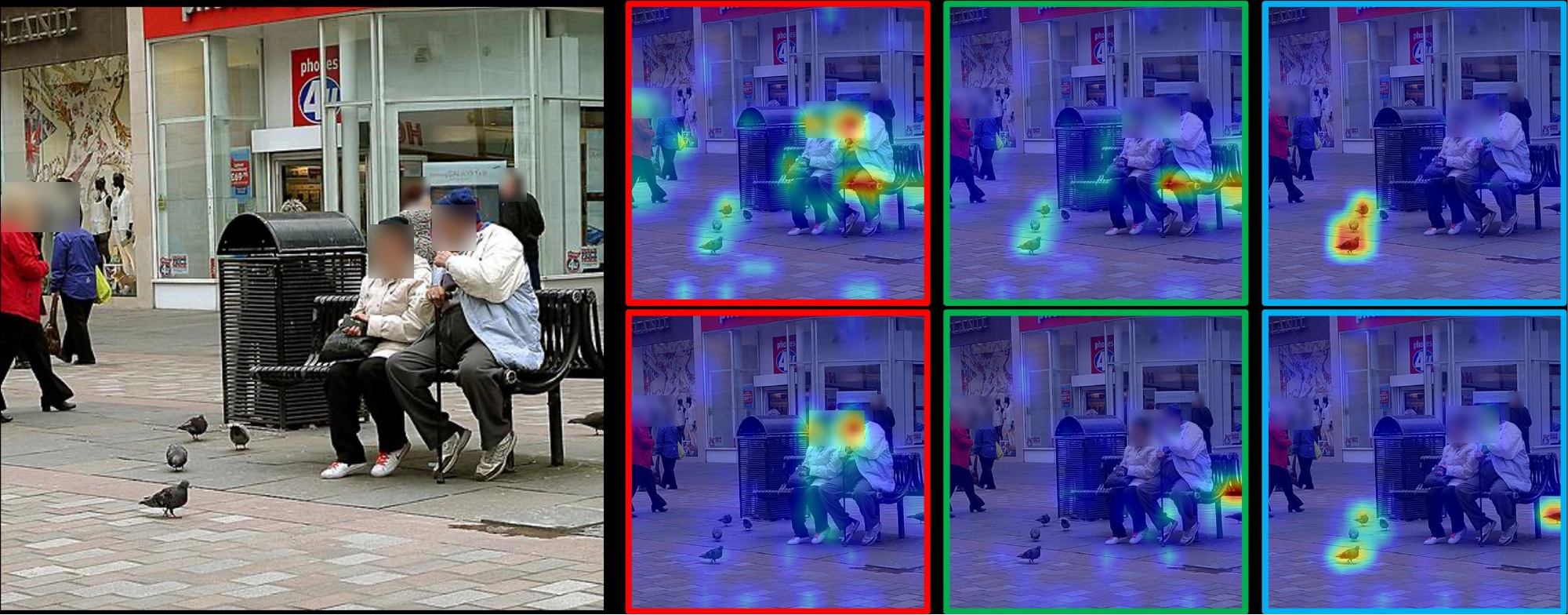}
        \subcaption{
            \textcolor[RGB]{255,0,0}{Two people sitting} on a \textcolor[RGB]{0,176,80}{bench} feeding \textcolor[RGB]{0,176,240}{birds}. %
        }
    \end{minipage}

    \begin{minipage}{\linewidth}
        \centering
        \includegraphics[width=\linewidth]{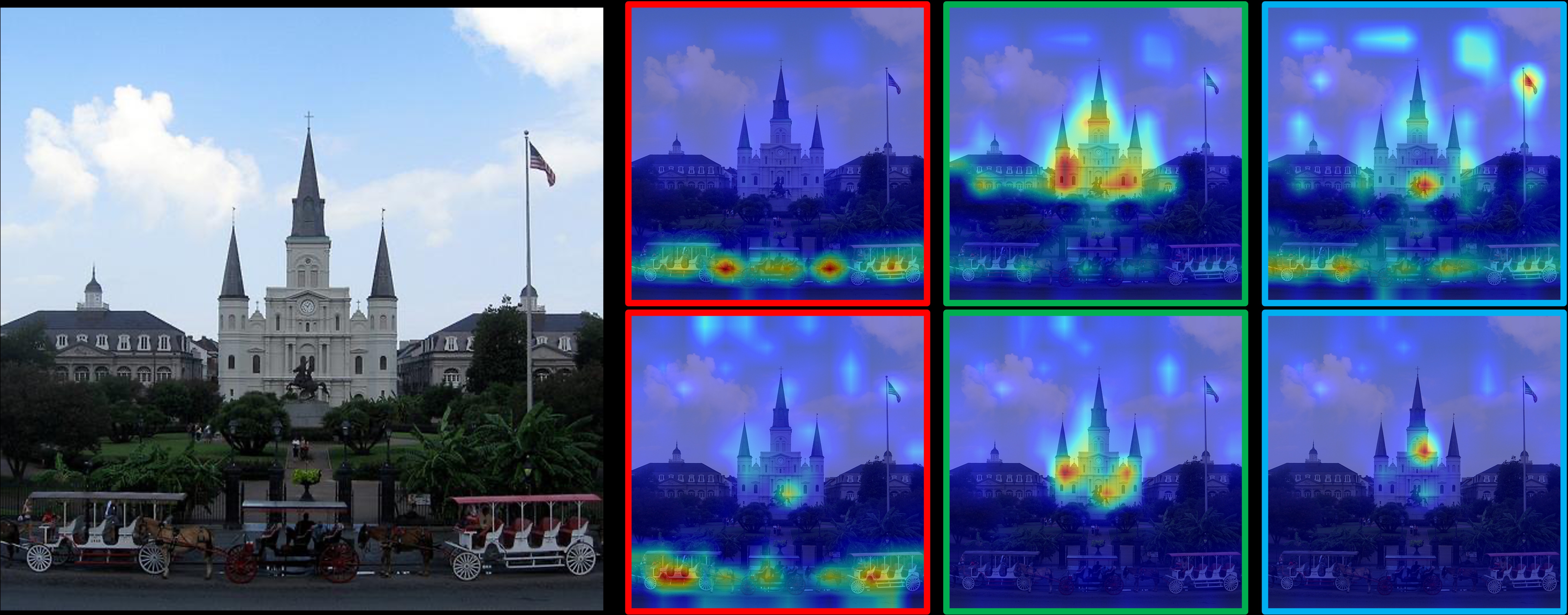}
        \subcaption{
            \textcolor[RGB]{255,0,0}{Three horse drawn carriages} in front of a \textcolor[RGB]{0,176,80}{castle} with a \textcolor[RGB]{0,176,240}{clock} on it.
        }
    \end{minipage}
\end{figure}

\end{document}